\documentclass{article}
\usepackage{ijcai16}

\usepackage{times}
\usepackage{bm}
\usepackage{amsfonts}
\usepackage{graphicx}
\usepackage{subfig}
\usepackage{algorithm}
\usepackage{algpseudocode}
\usepackage{multirow}
\usepackage{hyphenat}
\def\argmax{\mathop{\rm argmax}}

\title{Agreement-based Joint Training for \\ Bidirectional Attention-based Neural Machine Translation }
\author{Yong Cheng$^\#$,  Shiqi Shen$^\dagger$, Zhongjun He$^+$, Wei He$^+$, Hua Wu$^+$, Maosong Sun$^\dagger$, Yang Liu$^\dagger$ \thanks{Yang Liu is the corresponding author:  liuyang2011@ tsinghua. edu.cn.}\\
 $^\#$Institute for Interdisciplinary Information Sciences, Tsinghua University, Beijing, China  \\
 $^\dagger$State Key Laboratory of Intelligent Technology and Systems  \\
  Tsinghua National Laboratory for Information Science and Technology \\
 Department of Computer Science and Technology, Tsinghua University, Beijing, China\\
 $^+$Baidu Inc., Beijing, China  \\
}

\begin{document}

\maketitle

\begin{abstract}
The attentional mechanism has proven to be effective in improving end-to-end neural machine translation. However, due to the intricate structural divergence between natural languages, unidirectional attention-based models might only capture partial aspects of attentional regularities. We propose agreement-based joint training for bidirectional attention-based end-to-end neural machine translation. Instead of training source-to-target and target-to-source translation models independently, our approach encourages the two complementary models to agree on word alignment matrices on the same training data. Experiments on Chinese-English and English-French translation tasks show that agreement-based joint training significantly improves both alignment and translation quality over independent training.
\end{abstract}

\section{Introduction}
End-to-end neural machine translation (NMT) is a newly proposed paradigm for machine translation \cite{Kalchbrenner:13,Cho:14,Sutskever:14,Bahdanau:15}. Without explicitly modeling latent structures that are vital for conventional statistical machine translation (SMT) \cite{Brown:93,Koehn:03,Chiang:05}, NMT builds on an {\em encoder-decoder} framework: the encoder transforms a source-language sentence into a continuous-space representation, from which the decoder generates a target-language sentence.

While early NMT models encode a source sentence as a fixed-length vector, Bahadanau et al. \shortcite{Bahdanau:15} advocate the use of {\em attention} in NMT. They indicate that only parts of the source sentence have an effect on the target word being generated. In addition, the relevant parts often vary with different target words. Such an attentional mechanism has proven to be an effective technique in text generation tasks such as machine translation \cite{Bahdanau:15,Luong:15} and image caption generation \cite{Xu:15}.

However, due to the structural divergence between natural languages, modeling the correspondence between words in two languages still remains a major challenge for NMT, especially for distantly-related languages. For example, Luong et al. \shortcite{Luong:15} report that attention-based NMT lags behind the Berkeley aligner \cite{Liang:06} in terms of alignment error rate (AER) on the English-German data. One possible reason is that unidirectional attention-based NMT can only capture partial aspects of attentional regularities due to the non-isomorphism of natural languages.

In this work, we propose to introduce agreement-based learning \cite{Liang:06,Liang:07} into attention-based neural machine translation. The basic idea is to encourage source-to-target and target-to-source translation models to agree on word alignment on the same training data. This can be done by defining a new training objective that combines likelihoods in two directions as well as an agreement term that measures the consensus between word alignment matrices in two directions. Experiments on Chinese-English and English-French datasets show that our approach is capable of better accounting for attentional regularities and significantly improves alignment and translation quality over independent training.

\section{Background}
Given a source-language sentence $\mathbf{x}=\mathbf{x}_1, \dots, \mathbf{x}_m,\dots, \mathbf{x}_M$ that contains $M$ words and a target-language sentence $\mathbf{y}=\mathbf{y}_1,\dots, \mathbf{y}_{n},\dots, \mathbf{y}_{N}$ that contains $N$ words, end-to-end neural machine translation directly models the translation probability as a single, large neural network:
\begin{eqnarray}
P(\mathbf{y}|\mathbf{x};\bm{\theta})=\prod_{n=1}^{N}P(\mathbf{y}_n|\mathbf{x}, \mathbf{y}_{<n}; \bm{\theta}) \label{eq:prob}
\end{eqnarray}
where $\bm{\theta}$ is a set of model parameters and $\mathbf{y}_{<n}=\mathbf{y}_1,\dots, \mathbf{y}_{n-1}$ is a partial translation.

\begin{figure}[!t]
\centering
\includegraphics[width=0.4\textwidth]{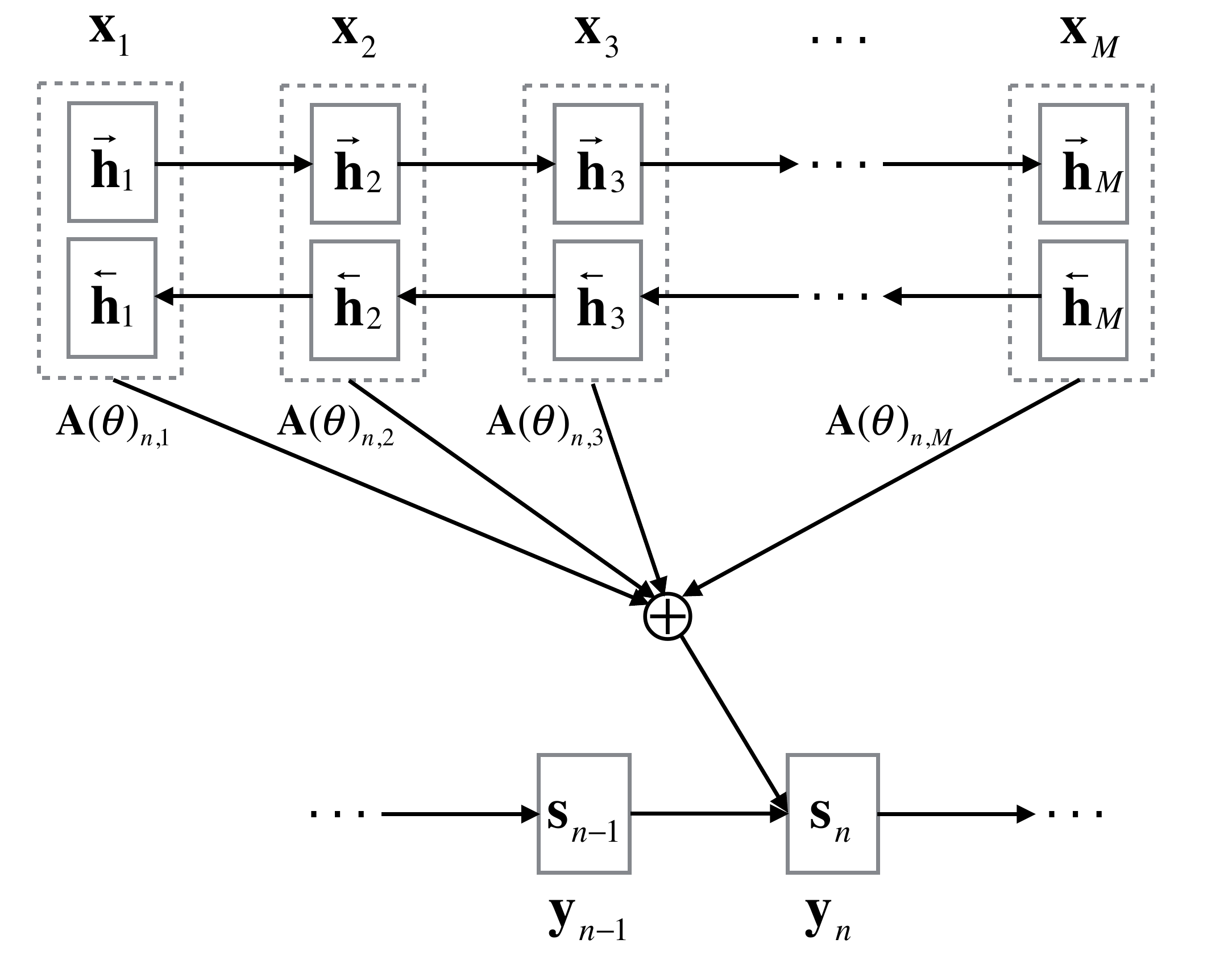}
\caption{The illustration of attention-based NMT. The decoder generates a target hidden state $\mathbf{s}_n$ and its corresponding target word $\mathbf{y}_n$ given a source sentence $\mathbf{x}$. A bidirectional RNN is used to concatenate the forward and backward states as the hidden states of source words.} \label{fig:attention}
\end{figure}

The encoder-decoder framework \cite{Kalchbrenner:13,Cho:14,Sutskever:14,Bahdanau:15} usually uses a recurrent neural network (RNN) to encode the source sentence into a sequence of hidden states $\mathbf{h}=\mathbf{h}_1,\dots,\mathbf{h}_m,\dots, \mathbf{h}_M$:
\begin{eqnarray}
\mathbf{h}_m = f(\mathbf{x}_m, \mathbf{h}_{m-1}, \bm{\theta})
\end{eqnarray}
where $\mathbf{h}_m$ is the hidden state of the $m$-th source word and $f(\cdot)$ is a non-linear function. Note that there are many ways to obtain the hidden states. For example, Bahdanau et al. \shortcite{Bahdanau:15} use a bidirectional RNN and concatenate the forward and backward states as the hidden state of a source word to capture both forward and backward contexts (see Figure \ref{fig:attention}).

\begin{figure*}[!t]

\centering
\begin{minipage}[t]{0.45\textwidth}
\centering
\includegraphics[width=1\textwidth]{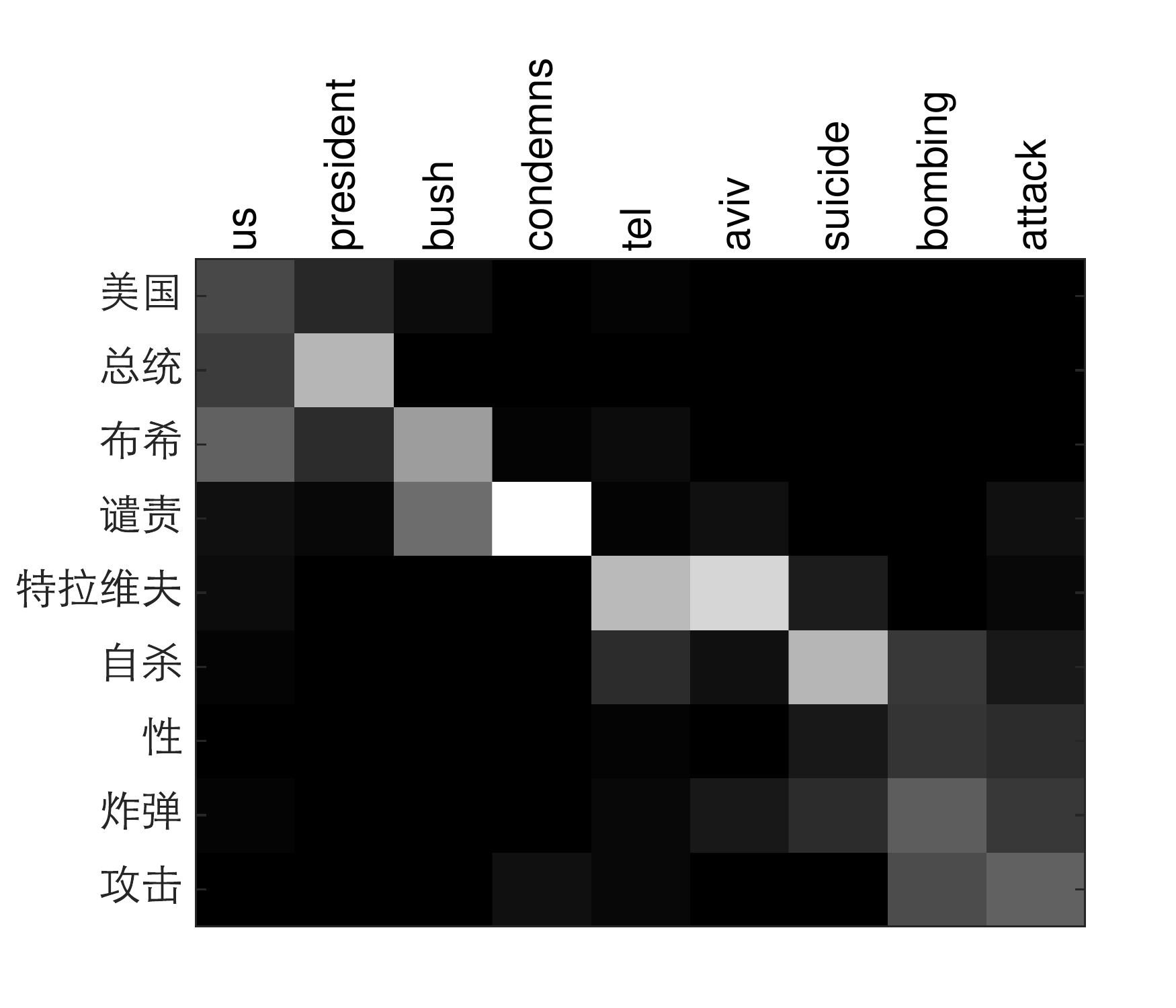}
\end{minipage}
\centering
\begin{minipage}[t]{0.45\textwidth}
\centering
\includegraphics[width=1\textwidth]{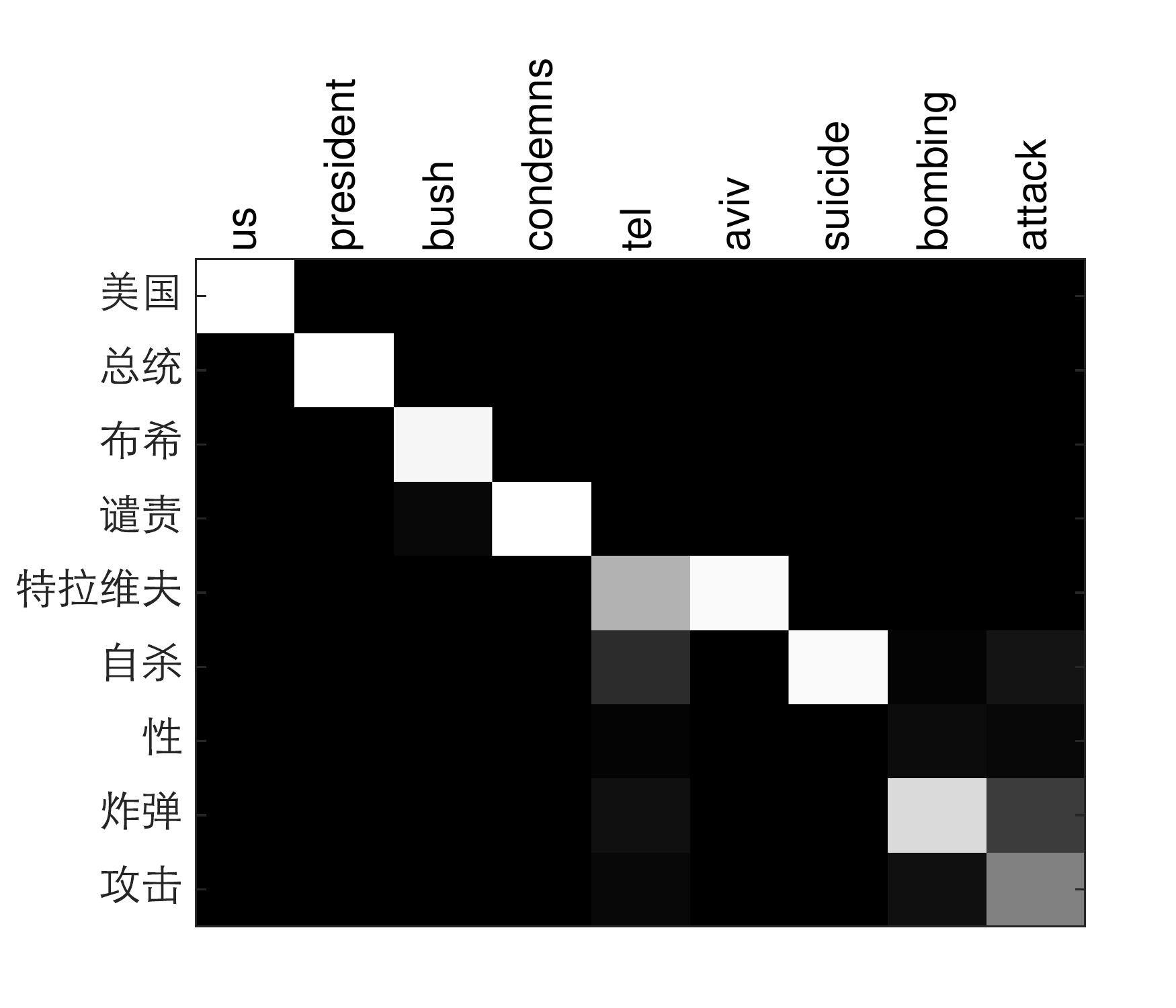}
\end{minipage}

\subfloat[independent training]{
\centering
\begin{minipage}[t]{0.45\textwidth}
\centering
\includegraphics[width=1\textwidth]{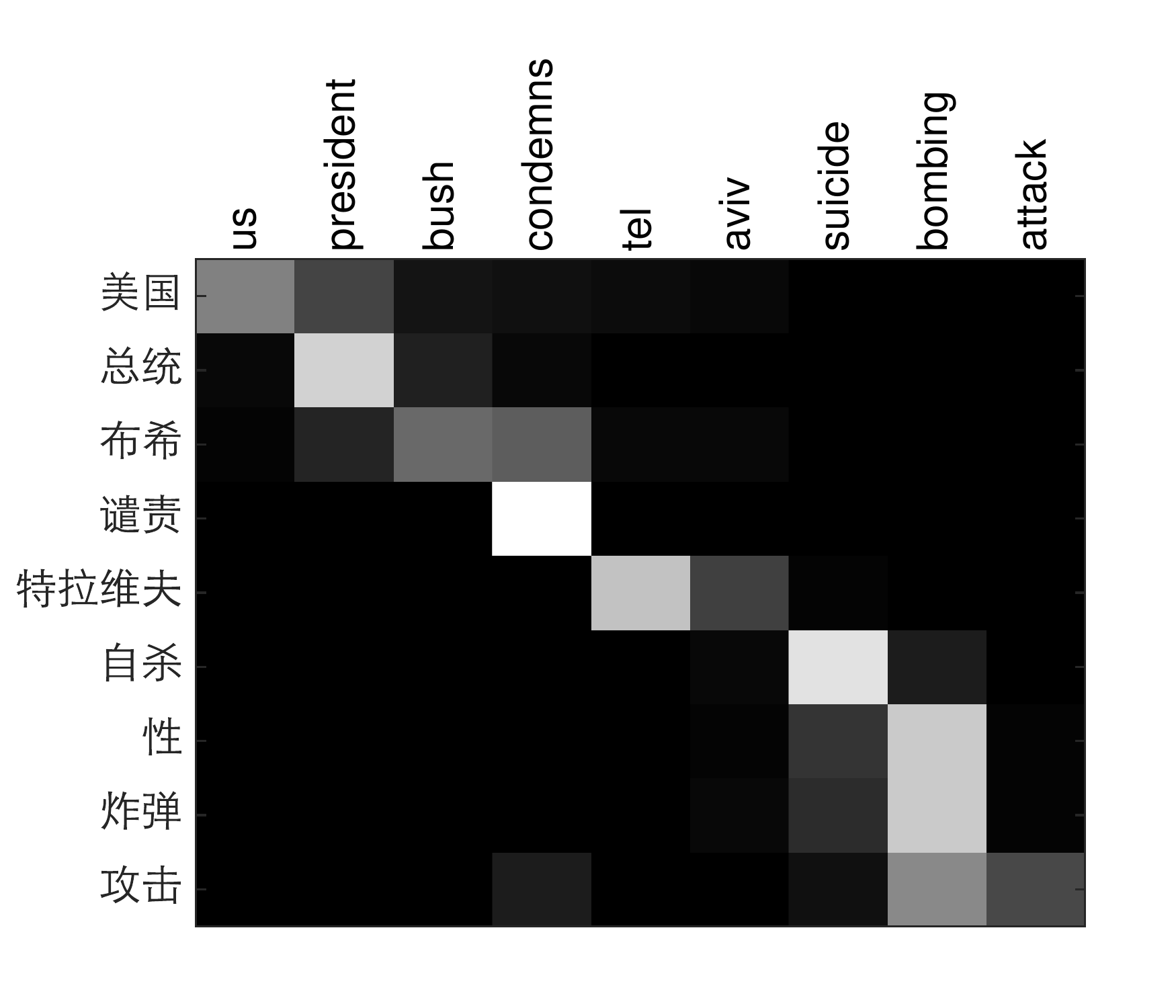}
\end{minipage}
}
\subfloat[joint training]{
\begin{minipage}[t]{0.45\textwidth}
\centering
\includegraphics[width=1\textwidth]{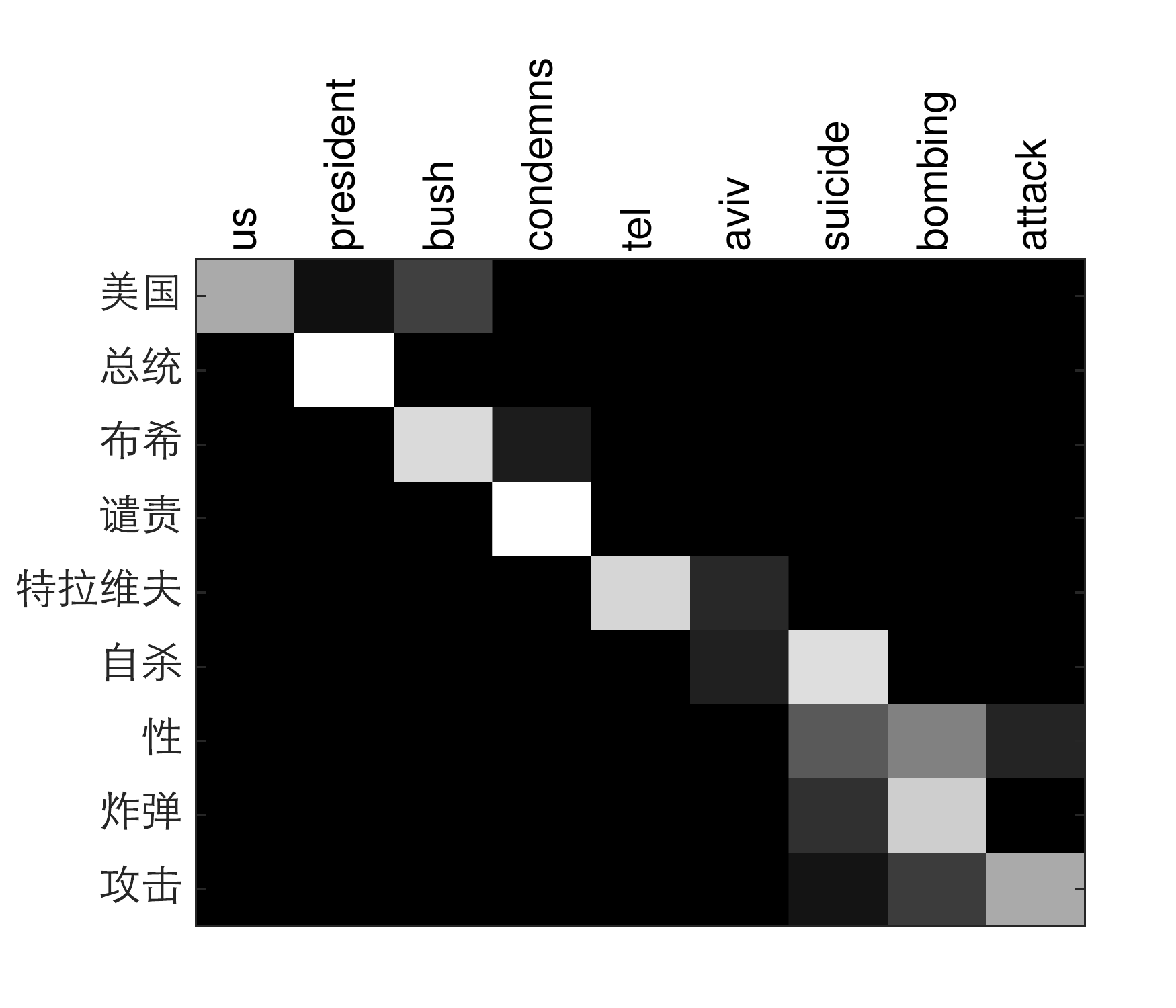}
\end{minipage}
}
\caption{Example alignments of (a) independent training and (b) joint training on a Chinese-English sentence pair. The first row shows Chinese-to-English alignments and the second row shows English-to-Chinese alignments. We find that the two unidirectional models are complementary and encouraging agreement leads to improved alignment accuracy.}\label{fig:align_example}
\end{figure*}

Bahdanau et al. \shortcite{Bahdanau:15} define the conditional probability in Eq. (\ref{eq:prob}) as
\begin{eqnarray}
P(\mathbf{y}_n|\mathbf{x},\mathbf{y}_{<n};\bm{\theta})=g(\mathbf{y}_{n-1},\mathbf{s}_n,\mathbf{c}_n, \bm{\theta})
\end{eqnarray}
where $g(\cdot)$ is a non-linear function, $\mathbf{s}_n$ is the hidden state corresponding to the $n$-th target word computed by
\begin{eqnarray}
\mathbf{s}_n = f(\mathbf{s}_{n-1}, \mathbf{y}_{n-1}, \mathbf{c}_{n}, \bm{\theta})
\end{eqnarray}
and $\mathbf{c}_n$ is a context vector for generating the $n$-th target word:
\begin{eqnarray}
\mathbf{c}_{n} = \sum_{m=1}^{M} \mathbf{A}(\bm{\theta})_{n,m} \mathbf{h}_m
\end{eqnarray}

We refer to $\mathbf{A}(\bm{\theta}) \in \mathbb{R}^{N\times M}$ as {\em alignment matrix}, in which an element $\mathbf{A}(\bm{\theta})_{n,m}$ reflects the contribution of the $m$-th source word $\mathbf{x}_m$ to generating the $n$-th target word $\mathbf{y}_n$: \footnote{We denote the alignment matrix as $\mathbf{A}(\bm{\theta})$ instead of $\alpha$ in \cite{Bahdanau:15} to emphasize that it is a function parameterized by $\bm{\theta}$ and differentiable. Although $\mathbf{s}_n$ and $\mathbf{c}_n$ also depend on $\bm{\theta}$, we omit the dependencies for simplicity.}
\begin{eqnarray}
\mathbf{A}(\bm{\theta})_{n,m} = \frac{\exp(a(\mathbf{s}_{n-1}, \mathbf{h}_m, \bm{\theta}))}{\sum_{m'=1}^{M} \exp(a(\mathbf{s}_{n-1}, \mathbf{h}_{m'}, \bm{\theta}))}
\end{eqnarray}
where $a(\mathbf{s}_{n-1}, \mathbf{h}_{m}, \bm{\theta})$ measures how well $\mathbf{x}_m$ and $\mathbf{y}_n$ are aligned. Note that word alignment is treated as a function parameterized by $\bm{\theta}$ instead of a latent variable in attention-based NMT.

Given a set of training examples $\{ \langle \mathbf{x}^{(s)}, \mathbf{y}^{(s)} \rangle \}_{s=1}^{S}$, the training algorithm aims to find the model parameters that maximize the likelihood of the training data:
\begin{eqnarray}
\bm{\theta}^{*} = \argmax_{\bm{\theta}}\Bigg\{ \sum_{s=1}^{S} \log P(\mathbf{y}^{(s)}|\mathbf{x}^{(s)}; \bm{\theta}) \Bigg\}
\end{eqnarray}

Although the introduction of attention has advanced the state-of-the-art of NMT, it is still challenging for attention-based NMT to capture the intricate structural divergence between natural languages. Figure \ref{fig:align_example}(a) shows the Chinese-to-English (upper) and English-to-Chinese (bottom) alignment matrices for the same sentence pair. Both the two independently trained models fail to correctly capture the gold-standard correspondence: while the Chinese-to-English alignment assigns wrong probabilities to ``us'' and ``bush'', the English-to-Chinese alignment makes incorrect predictions on ``condemns'' and ``bombing''.

Fortunately, although each model only captures partial aspects of the mapping between words in natural languages, the two models seem to be complementary: the Chinese-to-English alignment does well on ``condemns'' and the English-to-Chinese alignment assigns correct probabilities to ``us'' and ``bush''. Therefore, combining the two models can hopefully improve alignment and translation quality in both directions.

\section{Agreement-based Joint Training}
In this work, we propose to introduce agreement-based learning \cite{Liang:06,Liang:07} into attention-based neural machine translation. The central idea is to encourage the source-to-target and target-to-source models to agree on alignment matrices on the same training data. As shown in Figure \ref{fig:align_example}(b), agreement-based joint training is capable of removing unlikely attention and resulting in more concentrated and accurate alignment matrices in both directions.

More formally, we train both the source-to-target attention-based neural translation model $P(\mathbf{y}|\mathbf{x}; \overrightarrow{\bm{\theta}})$ and the target-to-source model $P(\mathbf{x}|\mathbf{y}; \overleftarrow{\bm{\theta}})$ on a set of training examples $\{ \langle \mathbf{x}^{(s)}, \mathbf{y}^{(s)} \rangle \}_{s=1}^{S}$, where $\overrightarrow{\bm{\theta}}$ and $\overleftarrow{\bm{\theta}}$ are model parameters in two directions, respectively. The new training objective is given by
\begin{eqnarray}
J(\overrightarrow{\bm{\theta}}, \overleftarrow{\bm{\theta}}) &=& \sum_{s=1}^{S} \log P(\mathbf{y}^{(s)}|\mathbf{x}^{(s)}; \overrightarrow{\bm{\theta}}) \nonumber \\
&& + \sum_{s=1}^{S} \log P(\mathbf{x}^{(s)}|\mathbf{y}^{(s)}; \overleftarrow{\bm{\theta}}) \nonumber \\
&& - \lambda \sum_{s=1}^{S}\Delta \Big(\mathbf{x}^{(s)}, \mathbf{y}^{(s)}, \overrightarrow{\mathbf{A}}^{(s)}(\overrightarrow{\bm{\theta}}), \overleftarrow{\mathbf{A}}^{(s)}(\overleftarrow{\bm{\theta}}) \Big) \nonumber \\
&&
\end{eqnarray}
where $\overrightarrow{\mathbf{A}}^{(s)}(\overrightarrow{\bm{\theta}})$ is the source-to-target alignment matrix for the $s$-th sentence pair, $\overleftarrow{\mathbf{A}}^{(s)}(\overleftarrow{\bm{\theta}})$ is the target-to-source alignment matrix for the same sentence pair, $\Delta(\cdot)$ is a loss function that measures the disagreement between two matrices, and $\lambda$ is a hyper-parameter that balances the preference between likelihood and agreement.

For simplicity, we omit the dependency on the sentence pair and simply write the loss function as $\Delta \big(\overrightarrow{\mathbf{A}}^{(s)}(\overrightarrow{\bm{\theta}}), \overleftarrow{\mathbf{A}}^{(s)}(\overleftarrow{\bm{\theta}}) \big)$. While there are many alternatives for quantifying disagreement, we use the following three types of loss functions in our experiments:
\begin{enumerate}
\item {\em Square of addition} (SOA): the square of the element-wise addition of corresponding matrix cells
\begin{eqnarray}
\Delta_{\mathrm{SOA}} \big(\overrightarrow{\mathbf{A}}^{(s)}(\overrightarrow{\bm{\theta}}), \overleftarrow{\mathbf{A}}^{(s)}(\overleftarrow{\bm{\theta}}) \big) \quad \quad \quad \quad \quad \quad \ \ \nonumber \\
= -\sum_{n=1}^{N}\sum_{m=1}^{M}\Big( \overrightarrow{\mathbf{A}}^{(s)}(\overrightarrow{\bm{\theta}})_{n,m} + \overleftarrow{\mathbf{A}}^{(s)}(\overleftarrow{\bm{\theta}})_{m,n}  \Big)^2
\end{eqnarray}
Intuitively, this loss function encourages to increase the sum of the alignment probabilities in two corresponding matrix cells.

\item {\em Square of subtraction} (SOS): the square of the element-wise subtraction of corresponding matrix cells
\begin{eqnarray}
\Delta_{\mathrm{SOS}} \big(\overrightarrow{\mathbf{A}}^{(s)}(\overrightarrow{\bm{\theta}}), \overleftarrow{\mathbf{A}}^{(s)}(\overleftarrow{\bm{\theta}}) \big) \quad \quad \quad \quad \quad \quad \ \ \nonumber \\
= \sum_{n=1}^{N}\sum_{m=1}^{M}\Big( \overrightarrow{\mathbf{A}}^{(s)}(\overrightarrow{\bm{\theta}})_{n,m} - \overleftarrow{\mathbf{A}}^{(s)}(\overleftarrow{\bm{\theta}})_{m,n}  \Big)^2
\end{eqnarray}
Derived from the symmetry constraint proposed by Ganchev et al. \shortcite{ganchev2010posterior}, this loss function encourages that an aligned pair of words share close or even equal alignment probabilities in both directions.

\item {\em Multiplication} (MUL): the element-wise multiplication of corresponding matrix cells
\begin{eqnarray}
\Delta_{\mathrm{MUL}} \big(\overrightarrow{\mathbf{A}}^{(s)}(\overrightarrow{\bm{\theta}}), \overleftarrow{\mathbf{A}}^{(s)}(\overleftarrow{\bm{\theta}}) \big) \quad \quad \quad \quad \quad \ \ \ \ \nonumber \\
= -\log \sum_{n=1}^{N}\sum_{m=1}^{M} \overrightarrow{\mathbf{A}}^{(s)}(\overrightarrow{\bm{\theta}})_{n,m} \times \overleftarrow{\mathbf{A}}^{(s)}(\overleftarrow{\bm{\theta}})_{m,n}
\end{eqnarray}

This loss function is inspired by the agreement term \cite{Liang:06} and model invertibility regularization \cite{Levinboim:15}.

\end{enumerate}

The decision rules for the two directions are given by
\begin{eqnarray}
\overrightarrow{\bm{\theta}}^{*} = \argmax_{\overrightarrow{\bm{\theta}}}\Bigg\{ \sum_{s=1}^{S} \log P(\mathbf{y}^{(s)}|\mathbf{x}^{(s)}; \overrightarrow{\bm{\theta}}) - \quad \quad \nonumber \\
\lambda \sum_{s=1}^{S} \Delta\big(\overrightarrow{\mathbf{A}}^{(s)}(\overrightarrow{\bm{\theta}}), \overleftarrow{\mathbf{A}}^{(s)}(\overleftarrow{\bm{\theta}})\big) \Bigg\} \\
\overleftarrow{\bm{\theta}}^{*} = \argmax_{\overleftarrow{\bm{\theta}}}\Bigg\{ \sum_{s=1}^{S} \log P(\mathbf{x}^{(s)}|\mathbf{y}^{(s)}; \overleftarrow{\bm{\theta}}) - \quad \quad \nonumber \\
\lambda \sum_{s=1}^{S} \Delta\big(\overrightarrow{\mathbf{A}}^{(s)}(\overrightarrow{\bm{\theta}}), \overleftarrow{\mathbf{A}}^{(s)}(\overleftarrow{\bm{\theta}})\big) \Bigg\}
\end{eqnarray}

Note that all the loss functions are differentiable with respect to model parameters. It is easy to extend the original training algorithm for attention-based NMT \cite{Bahdanau:15} to implement agreement-based joint training since the two translation models in two directions share the same training data.

\section{Experiments}

\subsection{Setup}
We evaluated our approach on Chinese-English and English-French machine translation tasks.

For Chinese-English, the training corpus from LDC consists of 2.56M sentence pairs with 67.53M Chinese words and 74.81M English words. We used the NIST 2006 dataset as the validation set for hyper-parameter optimization and model selection. The NIST 2002, 2003, 2004, 2005, and 2008 datasets were used as test sets. In the NIST Chinese-English datasets, each Chinese sentence has four reference English translations. To build English-Chinese validation and test sets, we simply ``reverse'' the Chinese-English datasets: the first English sentence in the four references as the source sentence and the Chinese sentence as the single reference translation.

For English-French, the training corpus from WMT 2014 consists of 12.07M sentence pairs with 303.88M English words and 348.24M French words. The concatenation of news-test-2012 and news-test-2013 was used as the validation set and news-test-2014 as the test set. Each English sentence has a single reference French translation. The French-English evaluation sets can be easily obtained by reversing the English-French datasets.

We compared our approach with two state-of-the-art SMT and NMT systems:
\begin{enumerate}
\item \textproc{Moses} \cite{Koehn:07}: a phrase-based SMT system;
\item \textproc{RNNsearch} \cite{Bahdanau:15}: an attention-based NMT system.
\end{enumerate}

For \textproc{Moses}, we used the parallel corpus to train the phrase-based translation model and the target-side part of the parallel corpus to train a 4-gram language model using the SRILM \cite{Stolcke:02}. We used the default system setting for both training and decoding.

For \textproc{RNNsearch}, we used the parallel corpus to train the attention-based NMT models. The vocabulary size is set to 30K for all languages. We follow Jean et al. \shortcite{Jean:15} to address the unknown word problem based on alignment matrices. Given an alignment matrix, it is possible to calculate the position of the source word to which is most likely to be aligned for each target word. After a source sentence is translated, each unknown word is translated from its corresponding source word. While Jean et al. \shortcite{Jean:15} use a bilingual dictionary generated by an off-the-shelf word aligner to translate unknown words, we use unigram phrases instead.

Our system simply extends \textproc{RNNsearch} by replacing independent training with agreement-based joint training. The encoder-decoder framework and the attentional mechanism remain unchanged. The hyper-parameter $\lambda$ that balances the preference between likelihood and agreement is set to 1.0 for Chinese-English and 2.0 for English-French. The training time of joint training is about 1.2 times longer than that of independent training for two directional models. We used the same unknown word post-processing technique as \textproc{RNNsearch} for our system.

\subsection{Comparison of Loss Functions}
\begin{table}[!t]
\centering
\begin{tabular}{l|c}
Loss & BLEU \\
\hline
$\Delta_{\mathrm{SOA}}$: square of addition & 31.26 \\
$\Delta_{\mathrm{SOS}}$: square of subtraction & 31.65 \\
$\Delta_{\mathrm{MUL}}$: multiplication & 32.65
\end{tabular}
\caption{Comparison of loss functions in terms of case-insensitive BLEU scores on the validation set for Chinese-to-English translation.} \label{table:loss}
\end{table}

\begin{table*}[!t]
\centering
\begin{tabular}{c|c|c||l|lllll}
System & Training & Direction & NIST06 & NIST02 & NIST03 & NIST04 & NIST05 & NIST08 \\
\hline \hline
\multirow{2}{*}{\textproc{Moses}} & \multirow{2}{*}{indep.} & C$\rightarrow$E& 32.48 & 32.69 & 32.39 & 33.62 & 30.23 & 25.17 \\
& & E$\rightarrow$C & 14.27 & 18.28 & 15.36 & 13.96 & 14.11 & 10.84 \\
\hline
\multirow{4}{*}{\textproc{RNNsearch}} &  \multirow{2}{*}{indep.} & C$\rightarrow$E & 30.74 & 35.16 & 33.75 & 34.63 & 31.74 & 23.63 \\
& & E$\rightarrow$C & 15.71 & 20.76 & 16.56 & 16.85 & 15.14 & 12.70 \\
\cline{2-9}
& \multirow{2}{*}{joint} & C$\rightarrow$E & 32.65$^{++}$ & 35.68$^{**+}$ & 34.79$^{**++}$ & 35.72$^{**++}$ & 32.98$^{**++}$  & 25.62$^{*++}$ \\
& & E$\rightarrow$C & 16.25$^{*++}$ & 21.70$^{**++}$ & 17.45$^{**++}$ & 16.98$^{**}$ & 15.70$^{**+}$ & 13.80$^{**++}$\\
\end{tabular}
\caption{Results on the Chinese-English translation task. \textproc{Moses} is a phrase-based statistical machine translation system. \textproc{RNNsearch} is an attention-based neural machine translation system. We introduce agreement-based joint training for bidirectional attention-based NMT. NIST06 is the validation set and NIST02-05, 08 are test sets. The BLEU scores are case-insensitive. ``*'': significantly better than \textproc{Moses} ($p<0.05$); ``**'': significantly better than \textproc{Moses} ($p<0.01$); ``+'': significantly better than \textproc{RNNsearch} with independent training ($p<0.05$); ``++'': significantly better than \textproc{RNNsearch} with independent training ($p<0.01$). We use the statistical significance test with paired bootstrap resampling \protect\cite{koehn:04}. } \label{table:ce_translation}
\end{table*}

We first compared the three loss functions as described in Section 3 on the validation set for Chinese-to-English translation. The evaluation metric is case-insensitive BLEU.

 As shown in Table \ref{table:loss}, the square of addition loss function (i.e., $\Delta_{\mathrm{SOA}}$) achieves the lowest BLEU among the three loss functions. This can be possibly attributed to the fact that a larger sum does not necessarily lead to increased agreement. For example, while 0.9 + 0.1 hardly agree, 0.2 + 0.2 perfectly does. Therefore, $\Delta_{\mathrm{SOA}}$ seems to be an inaccurate measure of agreement.

 The square of subtraction loss function (i.e, $\Delta_{\mathrm{SOS}}$) is capable of addressing the above problem by encouraging the training algorithm to minimize the difference between two probabilities: (0.2 - 0.2)$^2$ = 0. However, the loss function fails to distinguish between (0.9 - 0.9)$^2$ and (0.2 - 0.2)$^2$. Apparently, the former should be preferred because both models have high confidence in the matrix cell. It is unfavorable for two models agree on a matrix cell but both have very low confidence. Therefore, $\Delta_{\mathrm{SOS}}$ is perfect for measuring agreement but ignores confidence.

As the multiplication loss function (i.e., $\Delta_{\mathrm{MUL}}$) is able to take both agreement and confidence into account (e.g., 0.9 $\times$ 0.9 $>$ 0.2 $\times$ 0.2), it achieves significant improvements over $\Delta_{\mathrm{SOA}}$ and $\Delta_{\mathrm{SOS}}$. As a result, we use $\Delta_{\mathrm{MUL}}$ in the following experiments.

\subsection{Results on Chinese-English Translation}

Table \ref{table:ce_translation} shows the results on the Chinese-to-English (C $\rightarrow$ E) and English-to-Chinese (E $\rightarrow$ C) translation tasks. \footnote{The scores for E $\rightarrow$ C is much lower than C $\rightarrow$ E because BLEU is calculated at the word level rather than character level.} We find that \textproc{RNNsearch} generally outperforms \textproc{Moses} except for the C $\rightarrow$ E direction on the NIST08 test set, which confirms the effectiveness of attention-based NMT on distantly-related language pairs such as Chinese and English.

Agreement-based joint training further systematically improves the translation quality in both directions over independently training except for the E $\rightarrow$ C direction on the NIST04 test set.

\subsection{Results on Chinese-English Alignment}

\begin{table}[!t]
\centering
\begin{tabular}{c|ll}
Training & C $\rightarrow$ E & E $\rightarrow$ C \\
\hline
indep. & 54.64 & 52.49 \\
joint & 47.49$^{**}$ & 46.70$^{**}$
\end{tabular}
\caption{Results on the Chinese-English word alignment task. The evaluation metric is alignment error rate. ``**'': significantly better than \textproc{RNNsearch} with independent training ($p<0.01$).} \label{table:ce_alignment}
\end{table}

Table \ref{table:ce_alignment} shows the results on the Chinese-English word alignment task. We used the \textproc{TsinghuaAligner} evaluation dataset \cite{Liu:15} in which both the validation and test sets contain 450 manually-aligned Chinese-English sentence pairs. We follow Luong et al. \shortcite{Luong:15} to ``force-decode'' our jointly trained models to produce translations that match the references. Then, we extract only one-to-one alignments by selecting the source word with the highest alignment weight for each target word.

We find that agreement-based joint training significantly reduces alignment errors for both directions as compared with independent training. This suggests that introducing agreement does enable NMT to capture attention more accurately and thus lead to better translations. Figure \ref{fig:align_example}(b) shows example alignment matrices resulted from agreement-based joint training.

However, the error rates in Table \ref{table:ce_alignment} are still higher than conventional aligners that can achieve an AER around 30 on the same dataset. There is still room for improvement in attention accuracy.

\subsection{Analysis of Alignment Matrices}

\begin{table}[!t]
\centering
\begin{tabular}{ccc|cc}
Word & Type & Freq. & Indep. & Joint \\
\hline
to & preposition & high & 2.21 & 1.80 \\
and & conjunction & high & 2.21 & 1.60 \\
the & definite article & high & 1.96 & 1.56 \\
yesterday & noun & medium & 2.04 & 1.55 \\
actively & adverb & medium & 1.90 & 1.32 \\
festival & noun & medium & 1.55 & 0.85 \\
inspects & verb & low & 0.29 & 0.02 \\
rebellious & adjective & low & 0.29 & 0.02 \\
noticing & verb & low & 0.19 & 0.01
\end{tabular}
\caption{Comparison of independent and joint training in terms of average attention entropy (see Eq. (\ref{eq:aae})) on Chinese-to-English translation.} \label{table:aae}
\end{table}

\begin{table*}[!t]
\centering
\begin{tabular}{c|c|c||l|l}
System & Training & Direction & Dev. & Test \\
\hline \hline
\multirow{2}{*}{\textproc{Moses}} & \multirow{2}{*}{Indep.} & E$\rightarrow$F& 28.38 & 32.31 \\
& & F$\rightarrow$E& 28.52 & 30.93 \\
\hline
\multirow{4}{*}{\textproc{RNNsearch}} & \multirow{2}{*}{Indep.} & E$\rightarrow$F& 29.06  & 32.69 \\
& & F$\rightarrow$E& 28.32  &  29.99\\
\cline{2-5}
& \multirow{2}{*}{Joint} & E$\rightarrow$F& 29.86$^{**++}$ & 33.45$^{**++}$ \\
& & F$\rightarrow$E& 29.01$^{**++}$  & 31.51$^{**++}$\\
\end{tabular}
\caption{Results on the English-French translation task. The BLEU scores are case-insensitive. ``**'': significantly better than \textproc{Moses} ($p<0.01$);  ``++'': significantly better than \textproc{RNNsearch} with independent training ($p<0.01$).} \label{table:ef_translation}
\end{table*}

We observe that a target word is prone to connect to too many source words in the alignment matrices produced by independent training. For example, in the lower alignment matrix of Figure \ref{fig:align_example}(a), the third Chinese word ``buxi'' is aligned to three English words: ``president'', ``bush'', and ``condemns''. In addition, all the three alignment probabilities are relatively low. Similarly, four English words contribute to generating the last Chinese word ``gongji'': ``condemns'', ``suicide'', ``boming'', and ``attack''.

In contrast, agreement-based joint training leads to more concentrated alignment distributions. For example, in the lower alignment matrix of Figure \ref{fig:align_example}(b), the third Chinese word ``buxi'' is most likely to be aligned to ``bush''. Likewise, the attention to the last Chinese word ``gongji'' now mainly focuses on ``attack''.

To measure the degree of concentration of attention, we define the {\em attention entropy} of a target word in a sentence pair as follows:
\begin{eqnarray}
\mathrm{H}_{\mathbf{y}_n} = - \sum_{m=1}^{M} \mathbf{A}(\bm{\theta})_{n, m} \log \mathbf{A}(\bm{\theta})_{n, m}
\end{eqnarray}

Given a parallel corpus $D=\{ \langle \mathbf{x}^{(s)}, \mathbf{y}^{(s)} \rangle \}^{S}_{s=1}$, the {\em average attention entropy} is defined as
\begin{eqnarray}
\tilde{\mathrm{H}}_{y} = \frac{1}{c(y, D)} \sum_{s=1}^{S} \sum_{n=1}^{N} \delta(\mathbf{y}^{(s)}_n, y) \mathrm{H}_{\mathbf{y}^{(s)}_n} \label{eq:aae}
\end{eqnarray}
where $c(y, D)$ is the occurrence of a target word $y$ on the training corpus $D$:
\begin{eqnarray}
c(y,D) = \sum_{s=1}^{S} \sum_{n=1}^{N} \delta(\mathbf{y}^{(s)}_n, y)
\end{eqnarray}

Table \ref{table:aae} gives the average attention entropy of example words on the Chinese-to-English translation task. We find that the entropy generally goes downs with the decrease of word frequencies, which suggests that frequent target words tend to gain attention from multiple source words. Apparently, joint training leads to more concentrated attention than independent training. The gap seems to increase with the decrease of word frequencies.

\subsection{Results on English-to-French Translation}

Table \ref{table:ef_translation} gives the results on the English-French translation task. While \textproc{RNNsearch} with independent training achieves translation performance on par with \textproc{Moses}, agreement-based joint learning leads to significant improvements over both baselines. This suggests that our approach is general and can be applied to more language pairs.

\section{Related Work}

Our work is inspired by two lines of research: (1) attention-based NMT and (2) agreement-based learning.

\subsection{Attention-based Neural Machine Translation}
Bahdanau et al. \shortcite{Bahdanau:15} first introduce the attentional mechanism into neural machine translation to enable the decoder to focus on relevant parts of the source sentence during decoding. The attention mechanism allows a neural model to cope better with long sentences because it does not need to encode all the information of a source sentence into a fixed-length vector regardless of its length. In addition, the attentional mechanism allows us to look into the ``black box'' to gain insights on how NMT works from a linguistic perspective.

Luong et al. \shortcite{Luong:15a} propose two simple and effective attentional mechanisms for neural machine translation and compare various alignment functions. They show that attention-based NMT are superior to non-attentional models in translating names and long sentences.

After analyzing the alignment matrices generated by \textproc{RNNsearch} \cite{Bahdanau:15}, we find that modeling the structural divergence of natural languages is so challenging that unidirectional models can only capture part of alignment regularities. This finding inspires us to improve attention-based NMT by combining two unidirectional models. In this work, we only apply agreement-based joint learning to \textproc{RNNsearch}. As our approach does not assume specific network architectures, it is possible to apply it to the models proposed by Luong et al. \shortcite{Luong:15a}.

\subsection{Agreement-based Learning}
Liang et al. \shortcite{Liang:06} first introduce agreement-based learning into word alignment: encouraging asymmetric IBM models to agree on word alignment, which is a latent structure in word-based translation models \cite{Brown:93}. This strategy significantly improves alignment quality across many languages. They extend this idea to deal with more latent-variable models in grammar induction and predicting missing nucleotides in DNA sequences \cite{Liang:07}.

Liu et al. \shortcite{Liu:15a} propose generalized agreement for word alignment. The new general framework allows for arbitrary loss functions that measure the disagreement between asymmetric alignments. The loss functions can not only be defined between asymmetric alignments but also between alignments and other latent structures such as phrase segmentations.

In attention-based NMT, word alignment is treated as a parametrized function instead of a latent variable. This makes word alignment differentiable, which is important for training attention-based NMT models. Although alignment matrices in attention-based NMT are in principle ``symmetric'' as they allow for many-to-many soft alignments, we find that unidirectional modeling can only capture partial aspects of structure mapping. Our contribution is to adapt agreement-based learning into attentional NMT, which significantly improves both alignment and translation.

\section{Conclusion}
We have presented agreement-based joint training for bidirectional attention-based neural machine translation. By encouraging bidirectional models to agree on parametrized alignment matrices, joint learning achieves significant improvements in terms of alignment and translation quality over independent training. In the future, we plan to further validate the effectiveness of our approach on more language pairs.

\section*{Acknowledgements}

 This work was done while Yong Cheng and Shiqi Shen were visiting Baidu. This research is supported by the 973 Program (2014CB340501, 2014CB340505), the National Natural Science Foundation of China (No. 61522204, 61331013, 61361136003), 1000 Talent Plan grant, Tsinghua Initiative Research Program grants 20151080475 and a Google Faculty Research Award.

\bibliographystyle{named}
\bibliography{ijcai16}

\begin{thebibliography}{}

\bibitem[\protect\citeauthoryear{Bahdanau \bgroup \em et al.\egroup
  }{2015}]{Bahdanau:15}
Dzmitry Bahdanau, KyungHyun Cho, and Yoshua Bengio.
\newblock Neural machine translation by jointly learning to align and
  translate.
\newblock In {\em Proceedings of ICLR}, 2015.

\bibitem[\protect\citeauthoryear{Brown \bgroup \em et al.\egroup
  }{1993}]{Brown:93}
Peter~F. Brown, Stephen~A. Della~Pietra, Vincent~J. Della~Pietra, and Robert~L.
  Mercer.
\newblock The mathematics of statistical machine translation: Parameter
  estimation.
\newblock {\em Computational Linguisitics}, 1993.

\bibitem[\protect\citeauthoryear{Chiang}{2005}]{Chiang:05}
David Chiang.
\newblock A hierarchical phrase-based model for statistical machine
  translation.
\newblock In {\em Proceedings of ACL}, 2005.

\bibitem[\protect\citeauthoryear{Cho \bgroup \em et al.\egroup }{2014}]{Cho:14}
Kyunghyun Cho, Bart van Merrienboer, Caglar Gulcehre, Dzmitry Bahdanau, Fethi
  Bougares, Holger Schwenk, and Yoshua Bengio.
\newblock Learning phrase representations using rnn encoder-decoder for
  statistical machine translation.
\newblock In {\em Proceedings of EMNLP}, 2014.

\bibitem[\protect\citeauthoryear{Ganchev \bgroup \em et al.\egroup
  }{2010}]{ganchev2010posterior}
Kuzman Ganchev, Joao Gra{\c{c}}a, Jennifer Gillenwater, and Ben Taskar.
\newblock Posterior regularization for structured latent variable models.
\newblock {\em The Journal of Machine Learning Research}, 11:2001--2049, 2010.

\bibitem[\protect\citeauthoryear{Jean \bgroup \em et al.\egroup
  }{2015}]{Jean:15}
Sebastien Jean, Kyunghyun Cho, Roland Memisevic, and Yoshua Bengio.
\newblock On using very large target vocabulary for neural machine translation.
\newblock In {\em Proceedings of ACL}, 2015.

\bibitem[\protect\citeauthoryear{Kalchbrenner and
  Blunsom}{2013}]{Kalchbrenner:13}
Nal Kalchbrenner and Phil Blunsom.
\newblock Recurrent continuous translation models.
\newblock In {\em Proceedings of EMNLP}, 2013.

\bibitem[\protect\citeauthoryear{Koehn and Hoang}{2007}]{Koehn:07}
Philipp Koehn and Hieu Hoang.
\newblock Factored translation models.
\newblock In {\em Proceedings of EMNLP}, 2007.

\bibitem[\protect\citeauthoryear{Koehn \bgroup \em et al.\egroup
  }{2003}]{Koehn:03}
Philipp Koehn, Franz~J. Och, and Daniel Marcu.
\newblock Statistical phrase-based translation.
\newblock In {\em Proceedings of HLT-NAACL}, 2003.

\bibitem[\protect\citeauthoryear{Koehn}{2004}]{koehn:04}
Philipp Koehn.
\newblock Statistical significance tests for machine translation evaluation.
\newblock In {\em Proceedings of EMNLP}, 2004.

\bibitem[\protect\citeauthoryear{Levinboim \bgroup \em et al.\egroup
  }{2015}]{Levinboim:15}
Tomer Levinboim, Ashish Vaswani, and David Chiang.
\newblock Model invertibility regularization: Sequence alignment with or
  without parallel data.
\newblock In {\em Proceedings of NAACL}, 2015.

\bibitem[\protect\citeauthoryear{Liang \bgroup \em et al.\egroup
  }{2006}]{Liang:06}
Percy Liang, Ben Taskar, and Dan Klein.
\newblock Alignment by agreement.
\newblock In {\em Proceedings of NAACL}, 2006.

\bibitem[\protect\citeauthoryear{Liang \bgroup \em et al.\egroup
  }{2007}]{Liang:07}
Percy Liang, Dan Klein, and Michael~I. Jordan.
\newblock Agreement-based learning.
\newblock In {\em Proceedings of NIPS}, 2007.

\bibitem[\protect\citeauthoryear{Liu and Sun}{2015}]{Liu:15}
Yang Liu and Maosong Sun.
\newblock Contrastive unsupervised word alignment with non-local features.
\newblock In {\em Proceedings of AAAI}, 2015.

\bibitem[\protect\citeauthoryear{Liu \bgroup \em et al.\egroup
  }{2015}]{Liu:15a}
Chunyang Liu, Yang Liu, Huanbo Luan, Maosong Sun, and Heng Yu.
\newblock Generalized agreement for bidirectional word alignment.
\newblock In {\em Proceedings of EMNLP}, 2015.

\bibitem[\protect\citeauthoryear{Luong \bgroup \em et al.\egroup
  }{2015a}]{Luong:15a}
Minh-Thang Luong, Hieu Pham, and Christopher~D. Manning.
\newblock Effective approaches to attention-based neural machine translation.
\newblock In {\em Proceedings of EMNLP}, 2015.

\bibitem[\protect\citeauthoryear{Luong \bgroup \em et al.\egroup
  }{2015b}]{Luong:15}
Minh-Thang Luong, Ilya Sutskever, Quoc~V. Le, Oriol Vinyals, and Wojciech
  Zaremba.
\newblock Addressing the rare word problem in neural machine translation.
\newblock In {\em Proceedings of ACL}, 2015.

\bibitem[\protect\citeauthoryear{Stolcke}{2002}]{Stolcke:02}
Andreas Stolcke.
\newblock Srilm - an extensible language modeling toolkit.
\newblock In {\em Proceedings of ICSLP}, 2002.

\bibitem[\protect\citeauthoryear{Sutskever \bgroup \em et al.\egroup
  }{2014}]{Sutskever:14}
Ilya Sutskever, Oriol Vinyals, and Quoc~V. Le.
\newblock Sequence to sequence learning with neural networks.
\newblock In {\em Proceedings of NIPS}, 2014.

\bibitem[\protect\citeauthoryear{Xu \bgroup \em et al.\egroup }{2015}]{Xu:15}
Kelvin Xu, Jimmy~Lei Ba, Ryan Kiros, KyungHyun Cho, Aaron Courville, Ruslan
  Salakhutdinov, Richard~S. Zemel, and Yoshua Bengio.
\newblock Show, attend and tell: Neural image caption generation with visual
  attention.
\newblock In {\em Proceedings of ICML}, 2015.

\end{thebibliography}

\end{document}